\relax
\documentclass[letterpaper]{article}
\usepackage{ijcai17}
\usepackage{times}
\usepackage{helvet}
\usepackage{courier}
\usepackage{times}
\usepackage{multirow}
\usepackage{latexsym}
\usepackage{amssymb}
\usepackage{amsmath}
\usepackage{url}
\usepackage{breakurl}
\usepackage{graphicx}
\usepackage{subfig}
\usepackage{tikz}
\usepackage{pgfplots}
\usepackage{pgfplotstable}
\usepackage{adjustbox}

\DeclareMathOperator{\softmax}{\mathbf{softmax}}

\def\be{\mathbf{e}}

\def\bg{\textbf{g}}
\def\bh{\textbf{h}}

\def\bbr{\mathbf{r}}
\def\bt{\mathbf{t}}

\def\bv{\mathbf{v}}

\def\bx{\mathbf{x}}

\def\bz{\mathbf{z}}

\def\bW{\mathbf{W}}
\def\bU{\mathbf{U}}

\newenvironment{itemize*}%
  {\begin{itemize}%
    \setlength{\itemsep}{1pt}%
    \setlength{\parskip}{1pt}}%
  {\end{itemize}}
  \newenvironment{enumerate*}%
  {\begin{enumerate}%
    \setlength{\itemsep}{1pt}%
    \setlength{\parskip}{1pt}}%
  {\end{enumerate}}

\begin{document}

\title{Knowledge Graph Representation with\\ Jointly Structural and Textual Encoding}

\author{Jiacheng Xu, Kan Chen, Xipeng Qiu, Xuangjing Huang\\
School of Computer Science, Fudan University\\
\{jcxu13, kchen14, xpqiu, xjhuang\}@fudan.edu.cn
}

\maketitle

\begin{abstract}
The objective of knowledge graph embedding is to encode both entities and relations of knowledge graphs into continuous low-dimensional vector spaces. Previously, most works focused on symbolic representation of knowledge graph with structure information, which can not handle  new entities or entities with few facts well. In this paper, we propose a novel deep architecture to utilize both structural and textual information of entities. Specifically, we introduce three neural models to encode the valuable information from text description of entity, among which an attentive model can select related information as needed. Then, a gating mechanism is applied to integrate representations of structure and text into a unified architecture. Experiments show that our models outperform baseline by margin on link prediction and triplet classification tasks. Source codes of this paper will be available on Github.
\end{abstract}

\section{Introduction}

Knowledge graphs have been proved to benefit many artificial intelligence applications, such as relation extraction, question answering and so on. A knowledge graph consists of multi-relational data, having entities as nodes and relations as edges. An instance of fact is represented as a triplet (\textit{Head Entity, Relation, Tail Entity}), where the \textit{Relation} indicates a relationship between these two entities. In the past decades, great progress has been made in building large scale knowledge graphs, such as WordNet\cite{miller1995wordnet}, Freebase \cite{bollacker2008freebase}. However, most of them have been built either collaboratively or semi-automatically and as a result, they often suffer from incompleteness and sparseness.

The knowledge graph completion is to predict relations between entities based on existing triplets in a knowledge graph. Recently, a new powerful paradigm has been proposed to encode every element (entity or relation) of a knowledge graph into a low-dimensional vector space \cite{Bordes:2013,socher2013reasoning}. The representations of entities and relations are obtained by minimizing a global loss function involving all entities and relations. Therefore, we can do reasoning over knowledge graphs through algebraic computations.

\begin{figure*}[t]\centering
  \includegraphics[width=1\linewidth]{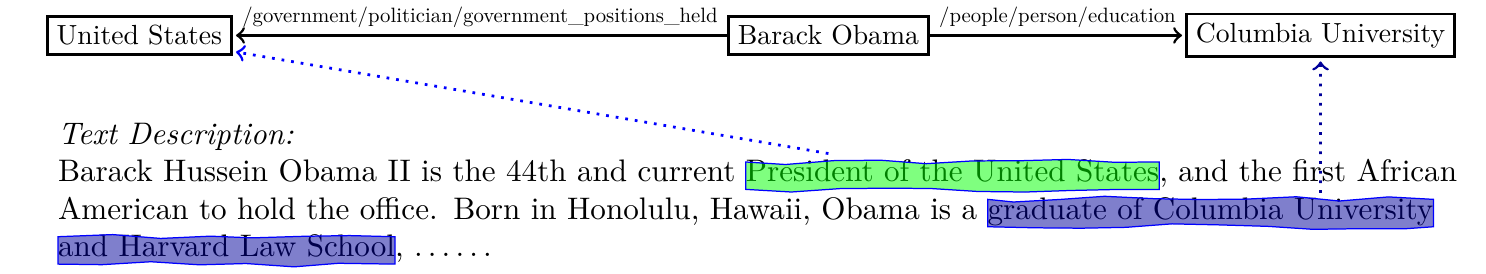}
  \caption{Example of entity description in Freebase.}\label{fig:example}
\end{figure*}

Although existing methods have good capability to learn knowledge graph embeddings, it remains challenging for entities with few or no facts \cite{ji2016knowledge}. To solve the issue of KB sparsity, many methods have been proposed to learn knowledge graph embeddings by utilizing related text information \cite{wang2014knowledge,zhong2015aligning,xie2016representation}.
These methods learn joint embedding of entities, relations, and words (or phrases, sentences) into the same vector space. However, there are still three problems to be solved. (1) The combination methods of the structural and textual representations are not well studied in these methods, in which two kinds of representations are merely aligned on word level or separate loss function. (2) The text description may represent an entity from various aspects, and various relations only focus on fractional aspects of the description. A good encoder should select the information from text in accordance with certain contexts of relations. Figure \ref{fig:example} illustrates the fact that not all information provided in its description are useful to predict the linked entities given a specific relation. (3) Intuitively, entities with many facts depend more on well-trained structured representation while those with few or no facts might be largely determined by text descriptions. A good representation should learn the most valuable information by balancing both sides.

In this paper, we propose a new deep architecture to learn the knowledge representation by utilizing the existing text descriptions of entities.  Specifically, we learn a joint representation of each entity from two information sources: one is structure information, and another is its text description. The joint representation is the combination of the structure and text representations with a gating mechanism. The gate decides how much information from the structure or text representation will carry over to the final joint representation.
In addition, we also introduce an attention mechanism to select the most related information from text description under different contexts.
Experimental results on link prediction and triplet classification show that our joint models can handle the sparsity problem well and outperform the baseline method on all metrics with a large margin.

Our contributions in this paper are summarized as follows.
\begin{enumerate}
  \item Unlike previous methods, we integrate the structure and text information of an entity into a joint representation, which can benefit the downstream applications.
  \item The gate mechanism can automatically find a balance between the structure and text information. For a low-frequency entity, the description will provide supplementary information for embedding, thus the issue of sparsity in knowledge base is settled properly.
  \item Given an entity, our attentive LSTM encoder can dynamically select the most related information from its text description according to different relations.
\end{enumerate}

\section{Knowledge Graph Embedding}

In this section, we briefly introduce the background knowledge about the knowledge graph embedding.

Knowledge graph embedding aims to model multi-relational data (entities and relations) into a continuous low-dimensional vector space. Given a pair of entities $(h,t)$ and their relation $r$, we can represent them with a triple $(h,r,t)$. A score function $f(h,r, t)$ is defined to model the correctness of the triple $(h,r,t)$, thus to distinguish whether two entities $h$ and $t$ are in a certain relationship $r$. $f(h,r, t)$ should be larger for a golden triplet $(h, r, t)$ that corresponds to a true fact in real world, otherwise $f(h,r, t)$ should be lower for an negative triplet.


The difference among the existing methods varies between linear \cite{Bordes:2013,Wang:2014} and nonlinear \cite{socher2013reasoning} score functions in the low-dimensional vector space.

Among these methods, TransE \cite{Bordes:2013} is a simple and effective approach, which learns the vector embeddings for both entities and relationships. Its basic idea is that the relationship between two entities is supposed to correspond to a translation between the embeddings of entities, that is, $\bh + \bbr \approx \bt$ when $(h,r,t)$ holds.

TransE's score function is defined as:
\begin{align}
f(h,r,t)) &= -\|\bh+\bbr-\bt\|_{2}^2
\end{align}
where $\bh,\bt,\bbr \in \mathbb{R}^d$ are embeddings of $h,t,r$ respectively, and satisfy $\|\bh\|^2_2=\|\bt\|^2_2=1$. The $\bh, \bbr, \bt$ are indexed by a lookup table respectively.

\section{Neural Text Encoding}

Given an entity in most of the existing knowledge bases, there is always an available corresponding text description with valuable semantic information for this entity, which can provide beneficial supplement for entity representation.

To encode the representation of a entity from its text description, we need to encode the variable-length sentence to a fixed-length vector. There are several kinds of neural models used in sentence modeling. These models generally consist of a projection layer that maps words, sub-word units or n-grams to vector representations (often trained beforehand with unsupervised methods), and then combine them with the different architectures of neural networks, such as neural bag-of-words (NBOW), recurrent neural network (RNN) \cite{Elman:1990,sutskever2014sequence,cho2014learning} and convolutional neural network (CNN) \cite{collobert2011natural,kalchbrenner2014convolutional}.

In this paper, we use three encoders (NBOW, LSTM and attentive LSTM) to model the text descriptions.

\subsection{Bag-of-Words Encoder}

A simple and intuitive method is the neural bag-of-words (NBOW) model, in which the representation of text can be generated by summing up its constituent word representations.

We denote the text description as word sequence $x_{1:n} = x_1,\cdots,x_n$, where $x_i$ is the word at position $i$. The NBOW encoder is
\begin{align}
\mathrm{enc_1}(x_{1:n}) = \sum_{i=1}^{n} \bx_i,
\end{align}
where $\bx_i \in \mathbb{R}^d$ is the word embedding of $x_i$.

\subsection{LSTM Encoder}

To address some of the modelling issues with NBOW, we consider using a bidirectional long short-term memory network (LSTM) \cite{schuster1997bidirectional,graves2005framewise} to model the text description.

LSTM was proposed by \cite{hochreiter1997long} to specifically address this issue of learning long-term dependencies \cite{bengio1994learning,hochreiter2001gradient,hochreiter1997long} in RNN. The LSTM maintains a separate memory cell inside it that updates and exposes its content only when deemed necessary.


Bidirectional LSTM (BLSTM) can be regarded as two separate LSTMs with different directions.
One LSTM models the text description from left to right, and another LSTM models text description from right to left respectively. We define the outputs of two LSTM at time step $i$ are $\overrightarrow \bz_i$ and $\overleftarrow \bz_i$ respectively.

The combined output of BLSTM at position $i$ is ${\bz_i} = \overrightarrow \bz_i \oplus \overleftarrow \bz_i$, where $\oplus$ denotes the concatenation operation.

The LSTM encoder combines all the outputs $\bz_i \in \mathbb{R}^d$ of BLSTM at different position.
\begin{align}
\mathrm{enc_2}(x_{1:n}) = \sum_{i=1}^{n} {\bz_i}.
\end{align}

\begin{figure}[t]\centering
  \includegraphics[width=1\linewidth]{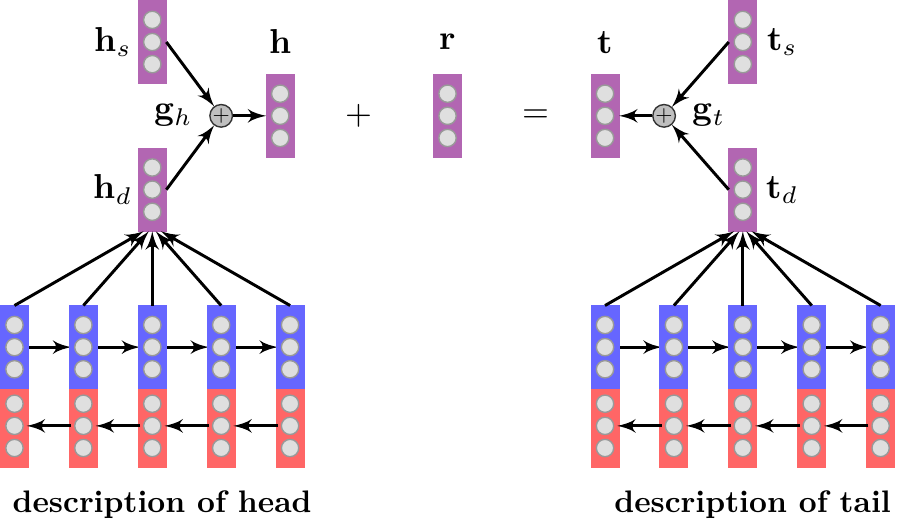}
  \caption{Our general architecture of jointly structural and textual encoding.}\label{fig:model}
\end{figure}

\subsection{Attentive LSTM Encoder}
While the LSTM encoder has richer capacity than NBOW, it produces the same representation for the entire text description regardless of its contexts.
However, the text description may present an entity from various aspects, and various relations only focus on fractional aspects of the description. This phenomenon also occurs in structure embedding for an entity \cite{Wang:2014,lin2015learning}.

Given a relation for an entity, not all of words/phrases in its text description are useful to model a specific fact. Some of them may be important for the given relation, but may be useless for other relations. Therefore, we introduce an attention mechanism \cite{bahdanau2014neural} to utilize an attention-based encoder that constructs contextual text encodings according to different relations.

For each position $i$ of the text description, the attention for a given relation $r$ is defined as $\alpha_i(r)$, which is
\begin{align}
e_i(r) &= \bv_a^T \tanh(\bW_a {\bz}_i + \bU_a \bbr), \\
\alpha_i(r)&=\softmax(e_i(r))\nonumber\\
&=\frac{\exp(e_i(r))}{\sum^{n}_{j=1} \exp(e_j(r))},
\end{align}
where $\bbr \in \mathbb{R}^d$ is the relation embedding; ${\bz}_i \in \mathbb{R}^d$ is the output of BLSTM at position $i$; $\bW_a,\bU_a \in \mathbb{R}^{d\times d}$ are parameters matrices; $\bv_a \in \mathbb{R}^{d}$ is a parameter vector.

The attention $\alpha_i(r)$ is interpreted as the degree to which the network attends to partial representation $\bz_{i}$ for given relation $r$.

The contextual encoding of text description can be formed by a weighted sum of the encoding $\bz_{i}$ with attention.
\begin{align}
\mathbf{enc_3}(x_{1:n};r) &= \sum_{i=1}^{n} \alpha_i(r) * \bz_i.
\end{align}

%
\section{Joint Structure and Text Encoder}

Since both the structure and text description provide valuable information for an entity , we wish to integrate all these information into a joint representation.


We propose a united model to learn a joint representation of both structure and text information. The whole model can be end-to-end trained.

For an entity $e$, we denote $\be_s$ to be its embedding of structure information, $\be_d$ to be encoding of its text descriptions. The main concern is how to combine $\be_s$ and $\be_d$.

To integrate two kinds of representations of entities, we use gating mechanism to decide how much the joint representation depends on structure or text.

The joint representation $\be$ is a linear interpolation between the $\be_s$ and $\be_d$.
\begin{align}
\be = \bg_e \odot \be_s + (1-\bg_e)\odot \be_d,
\end{align}
where $\bg_e$ is a gate to balance two sources information and its elements are in $[0,1]$, and $\odot$ is an element-wise multiplication. Intuitively, when the gate is close to 0, the joint representation is forced to ignore the structure information and is the text representation only.



\paragraph{Gate Strategy} We set $\bg_e$ to be a static vector, which means all the dimensions of $\be_s$ and $\be_d$ are summed by the different weights. We assign a static gate $\bg_e$ to each entity $e$. To constrain the value of each element is in $[0,1]$, we use logistic sigmoid function to compute the gate.
\begin{align}
\bg_e = \sigma(\tilde{\bg}_e),
\end{align}
where $\tilde{\bg}_e \in \mathbb{R}^d$ is real-value vector and stored in a lookup table. Once $\tilde{\bg}_e$ is learned on training data, it keeps unchanged during test.


\begin{table}[]
  \centering
\begin{adjustbox}{max width=.5\textwidth}
  \begin{tabular}{|c|c|c|c|c|c|}
    \hline
Dataset & \#Rel& \#Ent & \#Train & \#Valid & \#Test\\ \hline
FB15k & 1,345 & 14,951 & 483,142 & 50,000 & 59,071\\ \hline
WN18 & 18 & 40,493  & 141,442  & 5,000  & 5,000\\  \hline
  \end{tabular}
  \end{adjustbox}
  \caption{Statistics of datasets used in experiments.}\label{tb:datasets}
\end{table}

\paragraph{Score Function}
Following TransE, our final score function is defined as
\begin{align}
& f(h,r,t;d_h,d_t) = \|\bigl(\bg_h \odot \bh_s + (1-\bg_h)\odot \bh_d  \bigl) \nonumber\\
& \qquad  + \bbr - \bigl(\bg_t \odot \bt_s + (1-\bg_t)\odot \bt_d \bigl)\|_{2}^2,
\end{align}
where $\bg_h$ and $\bg_t$ are gates of head and tail respectively.


Figure \ref{fig:model} gives an illustration of our model.
To model the structure information better, $\bh_s$, $\bbr$, $\bt_s$ can be pre-trained with one of existing methods of knowledge graph embeddings, such as TransE.

\begin{table*}[bt]\setlength{\tabcolsep}{3pt}
\centering
\begin{tabular}{|l|*{4}{cc|}}\hline
 Datasets & \multicolumn{4}{|c|}{WN18} & \multicolumn{4}{|c|}{FB15K} \\\hline
\multirow{2}{*}{Metric} & \multicolumn{2}{|c|}{Mean Rank} & \multicolumn{2}{|c|}{Hits@10} & \multicolumn{2}{|c|}{Mean Rank} & \multicolumn{2}{|c|}{Hits@10}\\\cline{2-9}
& Raw & Filt  & Raw  & Filt  & Raw  & Filt  & Raw  & Filt\\\hline
Unstructured \cite{bordes2012joint} & 315 &  304 & 35.3  & 38.2 & 1,074  & 979  & 4.5 &  6.3\\
SME (linear) \cite{bordes2012joint} &  545  &  533 &  65.1  &  74.1  &  274  &  154  &  30.7  &  40.8\\
SME (Bilinear) \cite{bordes2012joint} &  526  &  509 &  54.7  &  61.3 &  284  &  158  &  31.3  &  41.3\\
TransH \cite{Wang:2014} & 318&  303 &75.4 &86.7 &212  &87&  45.7& 64.4\\
TransR \cite{lin2015learning} & 238&  225 &\textbf{79.8}  &\textbf{92.0} &198 &77 &48.2 &68.7\\
TransD \cite{ji2015knowledge} & 224&  212 &\textbf{79.6}  &\textbf{92.2} &194 &91 &\textbf{53.4}  &\textbf{77.3}\\
\hline
CNN+TransE \cite{xie2016representation} & - & - & - & - & 181 & 91 & 49.6 & 67.4 \\\hline
TransE (Baseline)  &  263  &  251 &  75.4  &  89.2 &  243  &  125  &  34.9  &  47.1\\
Jointly(CBOW) &142  & 130 & {78.5}& {89.9}& 183&  92& 48.9& 67.4  \\
Jointly(LSTM) & \textbf{117}  &\textbf{95}& \textbf{79.5}&  \textbf{91.6}&  179& 90 &49.3 &69.7\\
Jointly(A-LSTM) & 134&  123&  78.6& 90.9& \textbf{167} &  \textbf{73} & \textbf{52.9}&  \textbf{75.5}\\\hline
\end{tabular}
\caption{Results on link prediction.}\label{tb:res-lp}
\end{table*}

\subsection{Training}
We use the contrastive max-margin criterion \cite{Bordes:2013,socher2013reasoning} to train our model. Intuitively, the max-margin criterion provides an alternative to probabilistic, likelihood-based estimation methods by concentrating directly on the robustness of the decision boundary of a model \cite{taskar2005learning}.  The main idea is that each triplet  $(h,r,t)$ coming from the training corpus should receives a higher score than a triplet in which one of the elements is replaced with a random elements.

We assume that there are $n_t$ triplets in training set and denote the $i$th triplet by $(h_i, r_i, t_i),(i = 1, 2, \cdots ,n_t)$. Each triplet has a label $y_i$ to indicate the triplet is positive ($y_i = 1$) or negative ($y_i = 0$).

Then the golden and negative triplets are denoted by $\mathcal{D} = \{(h_j, r_j, t_j) | y_j = 1\}$ and $\mathcal{\hat{D}} = \{(h_j, r_j, t_j) | y_j = 0\}$, respectively. The positive example are the triplets from training dataset, and the negative examples are generated as follows: $ \mathcal{\hat{D}} = \{(h_l, r_k, t_k) | h_l \neq h_k \wedge y_k = 1\}\cup \{(h_k, r_k, t_l) | t_l \neq t_k \wedge y_k = 1\}\cup \{(h_k, r_l, t_k) | r_l \neq r_k \wedge y_k = 1\}$. The sampling strategy is Bernoulli distribution described in \cite{Wang:2014}. 

Let the set of all parameters be $\Theta$, we minimize the following objective:
\begin{gather}
J(\Theta)=\sum_{(h,r,t) \in \mathcal{D}}\sum_{( \hat{h},\hat{r},\hat{t}) \in \mathcal{\hat{D}}} \max \left(0,\gamma - \right. \nonumber\\
f( h,r,t)+f(\hat{h},\hat{r},\hat{t})\left.\right)+ \eta \|\Theta\|_2^2,
\end{gather}
where $\gamma > 0$ is a margin between golden triplets and negative triplets., $f(h, r, t)$ is the score function. We use the standard $L_2$ regularization of all the parameters, weighted by the hyperparameter $\eta$.


\section{Experiment}
In this section, we study the empirical performance of our proposed models on two benchmark tasks: triplet classification and link prediction.


\subsection{Datasets}

We use two popular knowledge bases: WordNet \cite{miller1995wordnet} and Freebase \cite{bollacker2008freebase} in this paper.  Specifically, we use WN18 (a subset of WordNet) \cite{bordes2014semantic} and FB15K (a subset of Freebase) \cite{Bordes:2013} since their text descriptions are easily publicly available.\footnote{\url{https://github.com/xrb92/DKRL}} Table \ref{tb:datasets}  lists statistics of the two datasets.

\begin{table*}[t]\setlength{\tabcolsep}{3pt}
\centering
\begin{tabular}{|l|*{2}{cccc|}}\hline
 Tasks & \multicolumn{4}{|c|}{Prediction Head (Hits@10)} & \multicolumn{4}{|c|}{Prediction Tail (Hits@10)} \\\hline
 Relation Category & 1-to-1  & 1-to-N  & N-to-1  & N-to-N  & 1-to-1  & 1-to-N  & N-to-1  & N-to-N  \\\hline
TransE (Baseline) & 43.7 & 65.7 & 18.2 & 47.2 & 43.7  & 19.7 & 66.7 & 50.0 \\
Jointly(CBOW) & 75.4 &  91.6 &  18.5 &  44.1& 75.2 &  24.6 &  92.2 &  52.3\\
Jointly(LSTM) &81.3  & 88.9  &18.8  & 45.2& 80.1 & 25.4 & 89.6 & 52.4\\
Jointly(A-LSTM) & \textbf{83.8} & \textbf{95.1} & \textbf{21.1} & \textbf{47.9}  & \textbf{83} &  \textbf{30.8} & \textbf{94.7} & \textbf{53.1}\\\hline
\end{tabular}
\caption{Detailed results by category of relationship on FB15K.}\label{tb:res-lp-detailed-fb15k}
\end{table*}
\subsection{Link Prediction}
Link prediction is a subtask of knowledge graph completion to complete a triplet $(h, r, t)$ with $h$ or $t$ missing, i.e., predict $t$ given $(h, r)$ or predict $h$ given $(r, t)$. Rather than requiring one best answer, this task emphasizes more on ranking a set of candidate entities from the knowledge graph.

Similar to \cite{Bordes:2013}, we use two measures as our evaluation metrics. (1) Mean Rank: the averaged rank of correct entities or relations; (2) Hits@p: the proportion of valid entities or relations ranked in top $p$ predictions. Here, we set $p=10$ for entities and $p=1$ for relations. A lower Mean Rank and a higher Hits@p should be achieved by a good embedding model. We call this evaluation setting ``Raw''. Since a false predicted triplet may also exist in knowledge graphs, it should be regard as a valid triplet. Hence, we should remove the false predicted triplets included in training, validation and test sets before ranking (except the test triplet of interest). We call this evaluation setting ``Filter''. The evaluation results are reported under these two settings.

\paragraph{Implementation}
We select the margin $\gamma$ among $\{1, 2\}$, the embedding dimension $d$ among $\{20, 50, 100\}$, the regularization $\eta$ among $\{0, 1E{-5}, 1E{-6}\}$, two learning rates $\lambda_s$ and $\lambda_t$ among $\{0.001, 0.01, 0.05\}$ to learn the parameters of structure and text encoding. The dissimilarity measure is set to either $L_1$ or $L_2$ distance.

In order to speed up the convergence and avoid overfitting, we initiate the structure embeddings of entity and relation with the results of TransE. The embedding of a word is initialized by averaging the linked entity embeddings whose description include this word. The rest parameters are initialized by randomly sampling from uniform distribution in $[-0.1, 0.1]$.


The final optimal configurations are:
$\gamma= 2$, $d=20$, $\eta=1E{-5}$, $\lambda_s = 0.01$, $\lambda_t = 0.1$, and $L_1$ distance on WN18; $\gamma=2$, $d=100$, $\eta=1E{-5}$, $\lambda_s = 0.01$, $\lambda_t = 0.05$, and $L_1$ distance on FB15K.

\paragraph{Results}

Experimental results on both WN18 and FB15k are shown in Table \ref{tb:res-lp}, where we use ``Jointly(CBOW)'', ``Jointly(LSTM)'' and ``Jointly(A-LSTM)'' to represent our jointly encoding models with CBOW, LSTM and attentive LSTM text encoders. Our baseline is TransE since that the score function of our models is based on TransE.

From the results, we observe that proposed models surpass the baseline, TransE, on all metrics, which indicates that knowledge representation can benefit greatly from text description.


On WN18, the reason why ``Jointly(A-LSTM)'' is slightly worse than ``Jointly(LSTM)'' is probably because the number of relations is limited.
Therefore, the attention mechanism does not have obvious advantage. On FB15K, ``Jointly(A-LSTM)'' achieves the best performance and is significantly higher than baseline methods on mean rank.

Although the Hits@10 of our models are worse than the best state-of-the-art method, TransD, it is worth noticing that the score function of our models is based on TransE, not TransD. Our models are compatible with other state-of-the-art knowledge embedding models. We believe that our model can be further improved by adopting the score functions of other state-of-the-art methods, such as TransD.

Besides, textual information  largely alleviates the issue of sparsity and our model achieves substantial improvement on Mean Rank comparing with TransD. However, textual information may slightly degrade the representation of frequent entities which have been well-trained. This may be another reason why our Hits@10 is worse than TransD which only utilizes structural information.


\paragraph{Detailed Results by Category of Relationship}
For the comparison of Hits@10 of different kinds of relations, we categorized the relationships according to the cardinalities of their head and tail arguments into four classes: 1-to-1, 1-to-many, many-to-1, many-to-many. Mapping properties of relations follows the same rules in \cite{Bordes:2013}.

Table \ref{tb:res-lp-detailed-fb15k} shows the detailed results by mapping properties of relations on FB15k. We can see that our models outperform baseline TransE in all types of relations (1-to-1, 1-to-N, N-to-1 and N-to-N), especially when (1) predicting ``1-to-1'' relations and (2) predicting the 1 side for ``1-to-N'' and ``N-to-1'' relations.

\begin{figure}[t!]\centering
  \includegraphics[width=1\linewidth]{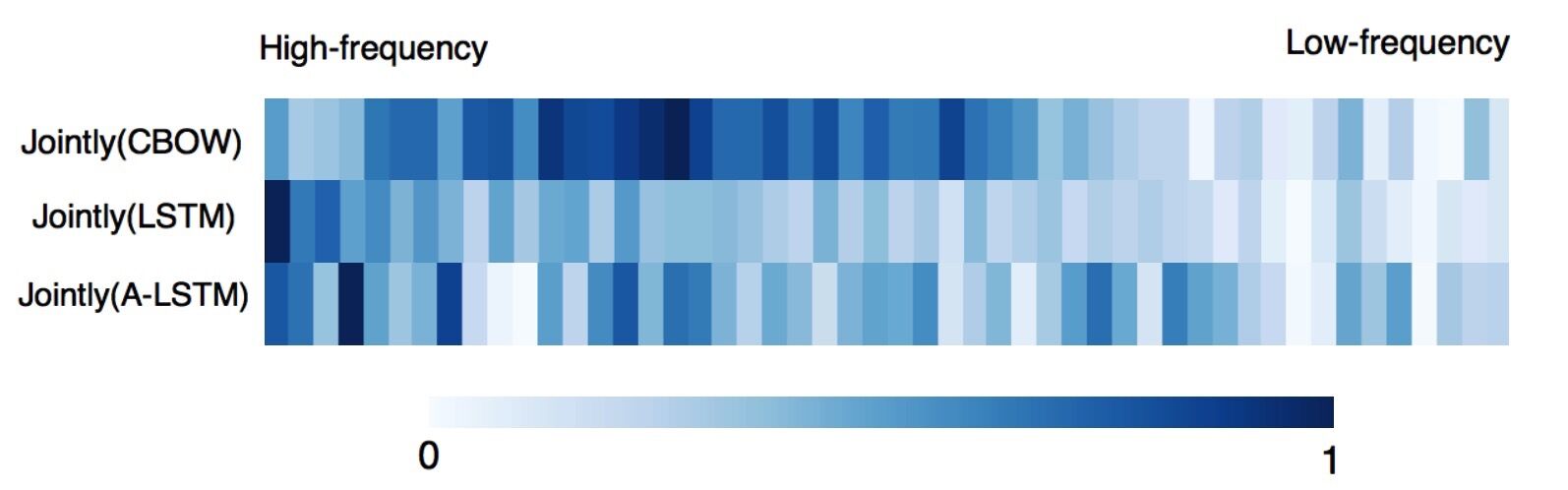}
  \caption{Visualization of gates with different entity frequencies on FB15K.}\label{fig:gate}
\end{figure}

\paragraph{Visualization of Gates}

 To get more insights into how the joint representation is influenced by the structure and text information. We observe the activations of gates, which control the balance between two sources of information, to understand the behavior of neurons. We sort the entities by their frequencies and divide them  into 50 equal-size groups of different frequencies, and average the values of all gates in each group.

Figure \ref{fig:gate} gives the average of gates in ten groups from high- to low-frequency. We observe that the text information play more important role for the low-frequency entities.

\subsection{Triplet Classification}
Triplet classification is a binary classification task, which aims to judge whether a given triplet $(h, r, t)$ is a correct fact or not.  Since our used test sets (WN18 and FB15K) only contain correct triplets, we construct negative triplets following the same setting used in \cite{socher2013reasoning}.

For triplets classification, we set a threshold $\delta_r$ for each relation $r$. $\delta_r$ is obtained by maximizing the classification accuracies on the valid set. For a given triplet $(h, r, t)$, if its score is larger than $\delta_r$, it will be classified as positive, otherwise negative.

\begin{table}[ht]
\centering
\begin{tabular}{|*{3}{c|}}\hline
 Datasets & WN18 & FB15K \\\hline
TransE & 92.9 &  79.8\\
TransH & - &  79.9\\
TransR & - &  82.1\\
CTransR & - & 84.3 \\
TransD & - &  88.0\\
TranSparse & - &  88.5\\
\hline
Jointly(NBOW) & 97.5 & 89.7\\
Jointly(LSTM) & 97.7 & 90.5\\
Jointly(A-LSTM) & \textbf{97.8}& \textbf{91.5} \\
\hline
\end{tabular}
\caption{Results on triplet classification.}\label{tb:res-tc}
\end{table}

\paragraph{Results}
Table \ref{tb:res-tc} shows the evaluation results of triplets classification. The results reveal that our joint encoding models is effective and also outperform the baseline method.

On WN18, ``Jointly(A-LSTM)'' achieves the best performance, and the ``Jointly(LSTM)'' is slightly worse than ``Jointly(A-LSTM)''. The reason is that the number of relations is relatively small. Therefore, the attention mechanism does not show obvious advantage.  On FB15K, the classification accuracy of  ``Jointly(A-LSTM)'' achieves 91.5\%, which is the best and significantly higher than that of state-of-the-art methods.

\section{Related Work}
Recently, it has gained lots of interests to jointly learn the embeddings of knowledge graph and text information. There are several methods using textual information to help KG representation learning.


\cite{socher2013reasoning} represent an entity as the average of its word embeddings in entity name, allowing the sharing of textual information located in similar entity names.

\cite{wang2014knowledge} jointly embed knowledge and text into the same space by aligning the entity name and its Wikipedia anchor, which brings promising improvements to the accuracy of predicting facts. \cite{zhong2015aligning}  extend the joint model and aligns knowledge and words in the entity descriptions.
However, these two works align the two kinds of embeddings on word level, which can lose some semantic information on phrase or sentence level.

\cite{zhang2015joint} also represent entities with entity names or the average of word embeddings in descriptions. However, their use of descriptions neglects word orders, and the use of entity names struggles with ambiguity.
\cite{xie2016representation} jointly learn knowledge graph embeddings with entity descriptions. They use continuous bag-of-words and convolutional neural network to encode semantics of entity descriptions. However, they separate the objective functions into two energy functions of structure-based and description-based representations.
\cite{Han:2016ua} embeds both entity and relation embeddings by taking KG and text into consideration using CNN.
To utilize both representations, they need further estimate an optimum weight coefficients to combine them together in the specific tasks.

Besides entity representation, there are also a lot of works \cite{lao2012reading,toutanova2015representing,neelakantan2015compositional}  to map textual relations and knowledge base relations to the same vector space and obtained substantial improvements.

While releasing the current paper we discovered a paper by \cite{Wu:2016wv} proposing a similar model with attention mechanism which is evaluated on link prediction and triplet classification. However, our work encodes text description as a whole without explicit segmentation of sentences, which breaks the order and coherence among sentences.

\section{Conclusion}
We propose a united representation for knowledge graph, utilizing both structure and text description information of the entities. Experiments show that our proposed jointly representation learning with gating mechanism is effective, which benefits to modeling the meaning of an entity.

In the future, we will consider the following research directions to improve our model:
\begin{enumerate*}
  \item Currently, our score function is based on TransE since the main focus of this work is how to integrate both structural and textual information. We believe our models can be further improved with the recently proposed knowledge graph embedding models.
  \item We will try to design dynamical gating strategy,which is estimated according to the context information.
  \item Intuitively, images of relations and entities may further improve the representation.
\end{enumerate*}


\bibliographystyle{named}
\bibliography{nlp,ours,jcxu}

\begin{thebibliography}{}

\bibitem[\protect\citeauthoryear{{Bahdanau} \bgroup \em et al.\egroup
  }{2014}]{bahdanau2014neural}
D.~{Bahdanau}, K.~{Cho}, and Y.~{Bengio}.
\newblock Neural machine translation by jointly learning to align and
  translate.
\newblock {\em ArXiv e-prints}, September 2014.

\bibitem[\protect\citeauthoryear{Bengio \bgroup \em et al.\egroup
  }{1994}]{bengio1994learning}
Yoshua Bengio, Patrice Simard, and Paolo Frasconi.
\newblock Learning long-term dependencies with gradient descent is difficult.
\newblock {\em Neural Networks, IEEE Transactions on}, 5(2):157--166, 1994.

\bibitem[\protect\citeauthoryear{Bollacker \bgroup \em et al.\egroup
  }{2008}]{bollacker2008freebase}
Kurt Bollacker, Colin Evans, Praveen Paritosh, Tim Sturge, and Jamie Taylor.
\newblock Freebase: a collaboratively created graph database for structuring
  human knowledge.
\newblock In {\em Proceedings of the 2008 ACM SIGMOD international conference
  on Management of data}, pages 1247--1250. ACM, 2008.

\bibitem[\protect\citeauthoryear{Bordes \bgroup \em et al.\egroup
  }{2012}]{bordes2012joint}
Antoine Bordes, Xavier Glorot, Jason Weston, and Yoshua Bengio.
\newblock Joint learning of words and meaning representations for open-text
  semantic parsing.
\newblock In {\em International Conference on Artificial Intelligence and
  Statistics}, pages 127--135, 2012.

\bibitem[\protect\citeauthoryear{Bordes \bgroup \em et al.\egroup
  }{2013}]{Bordes:2013}
Antoine Bordes, Nicolas Usunier, Alberto Garcia-Duran, Jason Weston, and Oksana
  Yakhnenko.
\newblock Translating embeddings for modeling multi-relational data.
\newblock In {\em NIPS}, 2013.

\bibitem[\protect\citeauthoryear{Bordes \bgroup \em et al.\egroup
  }{2014}]{bordes2014semantic}
Antoine Bordes, Xavier Glorot, Jason Weston, and Yoshua Bengio.
\newblock A semantic matching energy function for learning with
  multi-relational data.
\newblock {\em Machine Learning}, 94(2):233--259, 2014.

\bibitem[\protect\citeauthoryear{Cho \bgroup \em et al.\egroup
  }{2014}]{cho2014learning}
Kyunghyun Cho, Bart van Merrienboer, Caglar Gulcehre, Fethi Bougares, Holger
  Schwenk, and Yoshua Bengio.
\newblock Learning phrase representations using rnn encoder-decoder for
  statistical machine translation.
\newblock In {\em Proceedings of EMNLP}, 2014.

\bibitem[\protect\citeauthoryear{Collobert \bgroup \em et al.\egroup
  }{2011}]{collobert2011natural}
Ronan Collobert, Jason Weston, L{\'e}on Bottou, Michael Karlen, Koray
  Kavukcuoglu, and Pavel Kuksa.
\newblock Natural language processing (almost) from scratch.
\newblock {\em The Journal of Machine Learning Research}, 12:2493--2537, 2011.

\bibitem[\protect\citeauthoryear{Elman}{1990}]{Elman:1990}
Jeffrey~L Elman.
\newblock Finding structure in time.
\newblock {\em Cognitive science}, 14(2):179--211, 1990.

\bibitem[\protect\citeauthoryear{Graves and
  Schmidhuber}{2005}]{graves2005framewise}
Alex Graves and J{\"u}rgen Schmidhuber.
\newblock Framewise phoneme classification with bidirectional lstm and other
  neural network architectures.
\newblock {\em Neural Networks}, 18(5):602--610, 2005.

\bibitem[\protect\citeauthoryear{Han \bgroup \em et al.\egroup
  }{2016}]{Han:2016ua}
Xu~Han, Zhiyuan Liu, and Maosong Sun.
\newblock {Joint Representation Learning of Text and Knowledge for Knowledge
  Graph Completion}.
\newblock {\em arXiv.org}, November 2016.

\bibitem[\protect\citeauthoryear{Hochreiter and
  Schmidhuber}{1997}]{hochreiter1997long}
Sepp Hochreiter and J{\"u}rgen Schmidhuber.
\newblock Long short-term memory.
\newblock {\em Neural computation}, 9(8):1735--1780, 1997.

\bibitem[\protect\citeauthoryear{Hochreiter \bgroup \em et al.\egroup
  }{2001}]{hochreiter2001gradient}
Sepp Hochreiter, Yoshua Bengio, Paolo Frasconi, and J{\"u}rgen Schmidhuber.
\newblock Gradient flow in recurrent nets: the difficulty of learning long-term
  dependencies, 2001.

\bibitem[\protect\citeauthoryear{Ji \bgroup \em et al.\egroup
  }{2015}]{ji2015knowledge}
Guoliang Ji, Shizhu He, Liheng Xu, Kang Liu, and Jun Zhao.
\newblock Knowledge graph embedding via dynamic mapping matrix.
\newblock In {\em Proceedings of ACL}, pages 687--696, 2015.

\bibitem[\protect\citeauthoryear{Ji \bgroup \em et al.\egroup
  }{2016}]{ji2016knowledge}
Guoliang Ji, Kang Liu, Shizhu He, and Jun Zhao.
\newblock Knowledge graph completion with adaptive sparse transfer matrix.
\newblock In {\em AAAI}, 2016.

\bibitem[\protect\citeauthoryear{Kalchbrenner \bgroup \em et al.\egroup
  }{2014}]{kalchbrenner2014convolutional}
Nal Kalchbrenner, Edward Grefenstette, and Phil Blunsom.
\newblock A convolutional neural network for modelling sentences.
\newblock In {\em Proceedings of ACL}, 2014.

\bibitem[\protect\citeauthoryear{Lao \bgroup \em et al.\egroup
  }{2012}]{lao2012reading}
Ni~Lao, Amarnag Subramanya, Fernando Pereira, and William~W Cohen.
\newblock Reading the web with learned syntactic-semantic inference rules.
\newblock In {\em Proceedings of the 2012 Joint Conference on Empirical Methods
  in Natural Language Processing and Computational Natural Language Learning},
  pages 1017--1026. Association for Computational Linguistics, 2012.

\bibitem[\protect\citeauthoryear{Lin \bgroup \em et al.\egroup
  }{2015}]{lin2015learning}
Yankai Lin, Zhiyuan Liu, Maosong Sun, Yang Liu, and Xuan Zhu.
\newblock Learning entity and relation embeddings for knowledge graph
  completion.
\newblock In {\em AAAI}, 2015.

\bibitem[\protect\citeauthoryear{Miller}{1995}]{miller1995wordnet}
G.A. Miller.
\newblock Wordnet: a lexical database for english.
\newblock {\em Communications of the ACM}, 38(11):39--41, 1995.

\bibitem[\protect\citeauthoryear{Neelakantan \bgroup \em et al.\egroup
  }{2015}]{neelakantan2015compositional}
Arvind Neelakantan, Benjamin Roth, and Andrew McCallum.
\newblock Compositional vector space models for knowledge base completion.
\newblock {\em arXiv preprint arXiv:1504.06662}, 2015.

\bibitem[\protect\citeauthoryear{Schuster and
  Paliwal}{1997}]{schuster1997bidirectional}
Mike Schuster and Kuldip~K Paliwal.
\newblock Bidirectional recurrent neural networks.
\newblock {\em Signal Processing, IEEE Transactions on}, 45(11):2673--2681,
  1997.

\bibitem[\protect\citeauthoryear{Socher \bgroup \em et al.\egroup
  }{2013}]{socher2013reasoning}
Richard Socher, Danqi Chen, Christopher~D Manning, and Andrew Ng.
\newblock Reasoning with neural tensor networks for knowledge base completion.
\newblock In {\em Advances in Neural Information Processing Systems}, pages
  926--934, 2013.

\bibitem[\protect\citeauthoryear{Sutskever \bgroup \em et al.\egroup
  }{2014}]{sutskever2014sequence}
Ilya Sutskever, Oriol Vinyals, and Quoc~VV Le.
\newblock Sequence to sequence learning with neural networks.
\newblock In {\em Advances in Neural Information Processing Systems}, pages
  3104--3112, 2014.

\bibitem[\protect\citeauthoryear{Taskar \bgroup \em et al.\egroup
  }{2005}]{taskar2005learning}
Ben Taskar, Vassil Chatalbashev, Daphne Koller, and Carlos Guestrin.
\newblock Learning structured prediction models: A large margin approach.
\newblock In {\em Proceedings of the international conference on Machine
  learning}, 2005.

\bibitem[\protect\citeauthoryear{Toutanova \bgroup \em et al.\egroup
  }{2015}]{toutanova2015representing}
Kristina Toutanova, Danqi Chen, Patrick Pantel, Pallavi Choudhury, and Michael
  Gamon.
\newblock Representing text for joint embedding of text and knowledge bases.
\newblock In {\em Proceedings of the Conference on Empirical Methods in Natural
  Language Processing}, 2015.

\bibitem[\protect\citeauthoryear{Wang \bgroup \em et al.\egroup
  }{2014a}]{wang2014knowledge}
Zhen Wang, Jianwen Zhang, Jianlin Feng, and Zheng Chen.
\newblock Knowledge graph and text jointly embedding.
\newblock In {\em Proceedings of the Conference on Empirical Methods in Natural
  Language Processing (EMNLP)}, pages 1591--1601, 2014.

\bibitem[\protect\citeauthoryear{Wang \bgroup \em et al.\egroup
  }{2014b}]{Wang:2014}
Zhen Wang, Jianwen Zhang, Jianlin Feng, and Zheng Chen.
\newblock Knowledge graph embedding by translating on hyperplanes.
\newblock In {\em Proceedings of AAAI}, 2014.

\bibitem[\protect\citeauthoryear{Wu \bgroup \em et al.\egroup
  }{2016}]{Wu:2016wv}
Jiawei Wu, Ruobing Xie, Zhiyuan Liu, and Maosong Sun.
\newblock {Knowledge Representation via Joint Learning of Sequential Text and
  Knowledge Graphs}.
\newblock {\em arXiv.org}, September 2016.

\bibitem[\protect\citeauthoryear{Xie \bgroup \em et al.\egroup
  }{2016}]{xie2016representation}
Ruobing Xie, Zhiyuan Liu, Jia Jia, Huanbo Luan, and Maosong Sun.
\newblock Representation learning of knowledge graphs with entity descriptions.
\newblock In {\em Proceedings of IJCAI}, 2016.

\bibitem[\protect\citeauthoryear{Zhang \bgroup \em et al.\egroup
  }{2015}]{zhang2015joint}
Dongxu Zhang, Bin Yuan, Dong Wang, and Rong Liu.
\newblock Joint semantic relevance learning with text data and graph knowledge.
\newblock {\em ACL-IJCNLP 2015}, page~32, 2015.

\bibitem[\protect\citeauthoryear{Zhong \bgroup \em et al.\egroup
  }{2015}]{zhong2015aligning}
Huaping Zhong, Jianwen Zhang, Zhen Wang, Hai Wan, and Zheng Chen.
\newblock Aligning knowledge and text embeddings by entity descriptions.
\newblock In {\em Proceedings of EMNLP}, pages 267--272, 2015.

\end{thebibliography}

\end{document}